\pdfoutput=1

\documentclass[11pt]{article}

\usepackage{acl}
\usepackage{times}
\usepackage{latexsym}
\usepackage{textgreek}
\usepackage{mathtools}
\usepackage{amsmath}
\usepackage{txfonts}
\usepackage{xspace}
\usepackage{booktabs}
\usepackage{multirow}
\usepackage{enumitem} 
\usepackage{lipsum}
\usepackage{algorithm}        
\usepackage{pifont}
\usepackage{subcaption}
\usepackage[noend]{algpseudocode}
\usepackage{pgfplots}
\usepackage{pgfplotstable}
\usepackage{tikz}
\newcommand{\macrof}{$\mathrm{m}$-$\mathrm{F_1}$\xspace}
\newcommand{\microf}{$\mathrm{\muup}$-$\mathrm{F_1}$\xspace}

\usepackage{microtype}

\author{Ilias Chalkidis \\
Department of Computer Science, University of Copenhagen \\
\texttt{ilias.chalkidis[at]di.ku.dk}
}

\title{ChatGPT may Pass the Bar Exam soon, \\ but has a Long Way to Go for the LexGLUE benchmark}

\date{}

\begin{document}
\maketitle

\begin{abstract}

Following the hype around OpenAI's ChatGPT conversational agent, the last straw in the recent development of Large Language Models (LLMs) that demonstrate emergent unprecedented zero-shot capabilities, we audit the latest OpenAI's GPT-3.5 model, `\texttt{gpt-3.5-turbo}', the first available ChatGPT model, in the LexGLUE benchmark in a zero-shot fashion providing examples in a templated instruction-following format.
The results indicate that ChatGPT achieves an average micro-F1 score of 49.0\% across LexGLUE tasks, surpassing the baseline guessing rates. Notably, the model performs exceptionally well in some datasets, achieving micro-F1 scores of 62.8\% and 70.1\% in the ECtHR B and LEDGAR datasets, respectively. The code base and model predictions are available for review on \url{https://github.com/coastalcph/zeroshot_lexglue}.

\end{abstract}

\section{Introduction}

Recent advances in Large Language Models (LLMs)~\cite{brown-etal-2020-gpt3,chowdhery-etal-2022-palm}, also known as Foundation Models~\cite{bommasani-etal-2022-foundation}, have challenged the traditional supervised learning paradigm of fine-tuning by demonstrating emergent zero-shot Natural Language Understanding (NLU) capabilities~\cite{wei2022emergent} through scaling the model's size in billions of parameters~\cite{kaplan-etal-2020-scale}. 

OpenAI's latest conversational agent, ChatGPT\footnote{\url{https://chat.openai.com/chat}}~\cite{openai-2022-chatgpt}, a successor of InstructGPT~\cite{ouyang-etal-2022-instructgpt} -also known as GPT-3.5- models, is an instruction-following transformer-based language model, which has been further trained (\emph{aligned}) with reinforcement learning from human feedback (RLHF) \cite{christiano-etal-2017-drl}. ChatGPT demonstrates unprecedented emergent capabilities in zero-shot Question-Answering (QA) capabilities that cover common sense knowledge, but also extend to specialized domains such as problem solving, programming/debugging, and law, as presented by many users in the web. 

\begin{figure}[t]
    \centering
    \resizebox{\columnwidth}{!}{
    \includegraphics{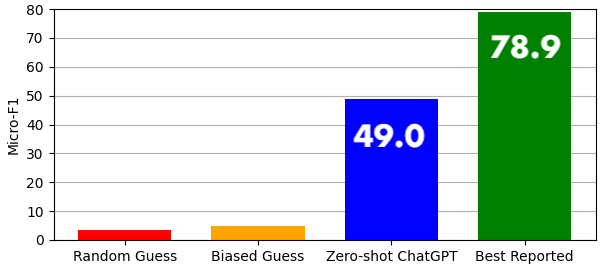}
    }
    \vspace{-4mm}
    \caption{Averaged performance on LexGLUE.}
    \label{fig:my_label}
    \vspace{-6mm}
\end{figure}

Recently, \citet{bommarito-bar-exams-2023} audited several variants of OpenAI's GPT 2/3/3.5 models in legal bar exam questions, and found that the most advanced -at the time- model (`\texttt{text-davinci-003}') achieves an accuracy of 50.3\% on a complete practice exam, significantly in excess of the 25\% baseline guessing rate, while it performs at a passing rate in two legal areas (Evidence and Torts). In a follow-up work, \citet{bommarito-etal-2023-gpt} assessed the model's performance in accounting certification exams, where the model significantly under-performs human capabilities with a correct rate of 14.4\%.

Following the work of \citeauthor{bommarito-bar-exams-2023}, we evaluate the latest OpenAI's GPT-3.5 model \cite{ouyang-etal-2022-instructgpt} (`\texttt{gpt-3.5-turbo}', v. March 2023), the first available ChatGPT, on legal text classification tasks from the LexGLUE \cite{chalkidis-etal-2022-lexglue} benchmark in a zero-shot fashion providing examples in a templated instruction-following format, similar to those used by \citet{chung-etal-2023-flant5}.  We find that ChatGPT achieves an average micro-F1 score of 49.0\% across LexGLUE tasks in a zero-shot setting, significantly in excess of the baseline guessing rates, while the model performs exceptionally well in some datasets achieving micro-F1 scores of 62.8\% and 70.1\% in the ECtHR B and LEDGAR datasets.

\section{Experiments}

\subsection{LexGLUE Datasets}

The LexGLUE~\cite{chalkidis-etal-2022-lexglue} benchmark comprises 7 legal text classification tasks: ECtHR A \& B~\cite{chalkidis-etal-2019-neural,chalkidis-etal-2021-paragraph}, SCOTUS~\cite{spaeth2020}, EURLEX~\cite{chalkidis-etal-2021-multieurlex}, LEDGAR~\cite{tuggener-etal-2020-ledgar}, UNFAIR-ToS~\cite{lippi-etal-2019-claudette}, and CaseHOLD~\cite{zhengguha2021}. We refer the readers to the work of \citeauthor{chalkidis-etal-2022-lexglue} for details on the tasks' formulation (input/output, labels).

To audit and evaluate ChatGPT, we sub-sample 1k samples from the test subset of each LexGLUE dataset (e.g., UNFAIR-ToS), and turn them into instructions using dataset-specific templates, which define how the input document, the classification question, and the available options (labels) are introduced. We opt to use and evaluate on a subset of the test sets aiming for a low compute budget up to ~\$50 (Table~\ref{tab:cost}).\footnote{In March 2023, the usage cost for `\texttt{gpt-3.5-turbo}' is \$0.002 / 1K tokens.} In 4 out of 7 datasets, the full test set comprises approx.~1k examples, except EURLEX with 5k, LEDGAR with approx.~10k, and CaseHOLD with approx.~4k examples.
We present an instruction-following example from the test set of UNFAIR-ToS:

\noindent\rule{6cm}{0.5pt}

\noindent\emph{\small \textbf{Given the following sentence from an online Term of Services:} ``if you are a resident of the european union (eu), please note that we offer this alternative  dispute resolution process, but we can not offer you the european commission dispute platform  as we do not have an establishment in the eu.''}\vspace{1mm}

\noindent\emph{\small \textbf{The sentence is unfair with respect to some of the following options:}}\vspace{1mm}

\noindent\emph{\small - Limitation of liability}\vspace{0.5mm}

\noindent\emph{\small - Unilateral termination}\vspace{0.5mm}

\noindent\emph{\small - Unilateral change}\vspace{0.5mm}

\noindent\emph{\small - Content removal}\vspace{0.5mm}

\noindent\emph{\small - Contract by using}\vspace{0.5mm}

\noindent\emph{\small - Choice of law}\vspace{0.5mm}

\noindent\emph{\small - Jurisdiction}\vspace{0.5mm}

\noindent\emph{\small - Arbitration}\vspace{0.5mm}

\noindent\emph{\small - None}\vspace{1mm}

\noindent\emph{\small \textbf{The relevant options are:} [None]}

\noindent\rule{6cm}{0.5pt}

\begin{table}[t]
    \centering
    \resizebox{0.95\columnwidth}{!}{
    \begin{tabular}{l|r|r}
\bf Dataset & \bf Avg. Length  & \bf Usage Cost \\ 
\midrule
ECtHR A    & 2.1k & \$4.43      \\ 
ECtHR B    & 2.1k & \$4.43      \\  
SCOTUS   & 3.6k & \$8.49      \\  
EURLEX    & 1.1k & \$6.15      \\  
LEDGAR   & 0.6k & \$1.34      \\  
UNFAIR-ToS & 0.1k & \$0.41      \\  
CaseHOLD   & 0.4k & \$0.99      \\  
\midrule
\multicolumn{2}{r|}{\bf Total} & \$26.24     \\ 
\bottomrule
\end{tabular}
}
\caption{Usage cost for processing 1k examples with the `\texttt{gpt-3.5-turbo}' model (v. March 2023). We report the averaged text length per instruction-following example measured in tokens.}
\label{tab:cost}
\vspace{-2mm}
\end{table}

\begin{table*}[t]
    \centering
    \resizebox{0.9\textwidth}{!}{
    \begin{tabular}{l|cc|cc||cc|cc|cc}
\bf \multirow{2}{*}{Dataset}    & \multicolumn{2}{c|}{\bf Random} & \multicolumn{2}{c||}{\bf Biased} & \multicolumn{2}{c|}{\bf Zero-shot} & \multicolumn{2}{c|}{\bf Few-shot} & \multicolumn{2}{c}{\bf Best} \\
 & \multicolumn{2}{c|}{\bf Guess} & \multicolumn{2}{c||}{\bf Guess} & \multicolumn{2}{c|}{\bf ChatGPT} & \multicolumn{2}{c|}{\bf ChatGPT} & \multicolumn{2}{c}{\bf Reported} \\
& \microf & \macrof & \microf & \macrof & \microf & \macrof & \microf & \macrof & \microf & \macrof \\
\midrule
ECtHR A    & 12.2 & 10.3 & 25.9 & 14.8 & 55.3 & 50.6     & \multicolumn{2}{c|}{n/a}                & 70.0 & 64.0                                 \\
ECtHR B    & 15.0 & 12.5 & 32.9 & 18.9 & 62.8 & 55.3     & \multicolumn{2}{c|}{n/a}                & 80.4 & 74.7                                       \\ 
SCOTUS     & 07.6 & 05.9 & 16.1 & 08.1 & 43.8 & 42.0     & \multicolumn{2}{c|}{n/a}                & 76.4 & 66.5                                       \\
EURLEX     & 05.1 & 03.8 & 12.0 & 04.8 & 32.5 & 21.1     & 24.8 & 13.2                             & 72.1 & 57.4                                 \\ 
LEDGAR     & 01.0 & 00.9 & 01.9 & 00.6 & 70.1 & 56.7     & 62.1 & 51.1                             & 88.2 & 83.0                                       \\ 
UNFAIR-ToS & 02.9 & 02.9 & 01.5 & 02.3 & 41.4 & 22.2     & 64.7 & 32.5                             & 96.0 & 83.0                                       \\ 
CaseHOLD   & 22.4 & 22.4 & 19.4 & 19.4 & 59.3 & 59.3     & \multicolumn{2}{c|}{n/a}                & 75.3 & 75.3                                       \\  
\midrule
Average    & 03.8 & 03.3 & 04.8 & 02.7  & 49.0 & 37.1 & \multicolumn{2}{c|}{n/a} & 78.9 & 70.8 \\
\bottomrule
    \end{tabular}
    }
    \vspace{-2mm}
    \caption{Test results across LexGLUE tasks. ChatGPT refers to the `\texttt{gpt-3.5-turbo}' GPT model (v. March 2023). We cannot evaluate the few-shot setting in several datasets (marked with n/a), since the documents are really long (2-4k words), and in many cases the documents have already been truncated up to 4k (incl. the template).}
    \label{tab:my_label}
    \vspace{-3mm}
\end{table*}

\subsection{Experimental Setup}

We evaluate the latest released OpenAI's `\texttt{gpt-3.5-turbo}' GPT model (v. March 2023), the first available ChatGPT model, using the \texttt{ChatCompletion} API\footnote{\url{https://platform.openai.com/docs/guides/chat}} providing one instruction-following example at a time as input with a generation limit up to 100 tokens -which is adequate to generate multiple label descriptors, if needed-. We perform an automated evaluation, where we primarily consider exact matches, i.e., the exact gold label descriptor is part of ChatGPT's generated answer. In case there are not any exact matches, we consider cosine similarity between the SentenceBERT~\cite{reimers-gurevych-2019-sentence} embeddings of all label descriptors and the generated answer, and assign the label with the highest similarity score. We report micro- (\microf), and macro-averaged F1 (\macrof) scores following \citet{chalkidis-etal-2022-lexglue}.

\subsection{Results \& Discussion}

In Table~\ref{tab:my_label}, we present the results across 5 different settings: (a) \emph{Random}, a random guess baseline, (b)  \emph{Biased}, a random guess baseline that considers the training distribution, (c) \emph{Zero-shot ChatGPT}, where the GPT model has no access to any training instances, (d) \emph{Few-shot ChatGPT}, where the GPT model has access to 8 training instances,\footnote{We evaluate few-shot prompting only on the datasets, where the input documents are not exceptionally long (2-4k), since ChatGPT has a processing limit of 4k tokens.} and (e) \emph{Best Reported} by \citet{chalkidis-etal-2022-lexglue} based on LegalBERT~\cite{chalkidis-etal-2020-legal} models fine-tuned on each dataset's training set.

We observe that the ChatGPT model is significantly outperformed by the supervized (fine-tuned) much smaller models in both settings (zero-shot, and few-shot). Nonetheless, the results of ChatGPT are clearly surpassing the random guess baselines, while in many cases the model performs exceptionally well even in the zero-shot setting, achieving micro-F1 (\microf) scores of 62.8\% and 70.1\% in the ECtHR B and LEDGAR datasets. 

We also observe that in the few-shot setting, the model benefits from a limited number of the task demonstrations ($K=8$) when the label set ($L$) is small ($K \approx L$), e.g., $L=9$ labels in UNFAIR-ToS, leading to a substantial performance improvement. In contrast, the performance deteriorates in both tasks (EURLEX, LEDGAR) with large label sets ($K=100$). We hypothesize that in the latter case, the limited number of demonstrations restricted to a very small number of labels ($K \ll L$), may bias the model to generate labels from the K-shot restricted set of labels.

Based on the aforementioned results, it is clear that the latest OpenAI's GPT-3.5 model, ChatGPT (`\texttt{gpt-3.5-turbo}'), has a non-trivial degree of legal knowledge; which is more vibrant is specific topic classification tasks (ECtHR B and LEDGAR). Nonetheless, much smaller fine-tuned models are still significantly better, and should be preferred in similar practical applications.

\section{Limitations}

Our findings are partly restricted to the developed instruction-following templates, since we do not perform any prompt engineering/tuning to restrict the computational budget. In Table~\ref{tab:ecthr_temps}, we present some additional results for ECtHR B with alternative (improved) instruction templates: \emph{Template \#1} with a basic question framing and simple label descriptors (e.g., ``Article 2''), \emph{Template \#2} with improved question framing, and \emph{Template \#3}, where we use the improved question framing and also provide more informative label descriptors (e.g., ``Article 2 - Right to life''). It is clear that prompt engineering can lead to minor performance improvements. Nonetheless, our code can be easily updated to run similar experiments with a different set of instruction-following templates that may be more suitable for the LexGLUE tasks.\footnote{\url{https://github.com/coastalcph/zeroshot_lexglue}}
Similarly, we do not experiment with different values for the sampling temperature, and use a fixed rate of 0 to get deterministic responses (predictions).

\begin{table}[t]
    \centering
    \resizebox{0.85\columnwidth}{!}{
    \begin{tabular}{l|cc}
    \bf Instruction Template & \microf & \macrof \\
    \midrule
    Template \#1 & 62.8 & 55.3 \\
     Template \#2 & \underline{64.0} & \underline{58.2} \\
     Template \#3 & 63.8 & 57.3 \\
     \bottomrule
    \end{tabular}
    }
    \caption{Alternative instruction templates in ECtHR B.}
    \label{tab:ecthr_temps}
    \vspace{-4mm}
\end{table}

We also do not experiment with other available non-conversational GTP-3.5 models (e.g., `\texttt{text-davinci-003}'), since the computation cost is 10$\times$ higher, i.e., the total cost would be approx.~\$500. We believe that this is not an important issue, since in the OpenAI's website, it is mentioned that: ``\emph{The performance of gpt-3.5-turbo is on par with Instruct Davinci} [officially named `\texttt{text-davinci-003}'].'', while also in the model's description page, they mention: \emph{``We recommend using gpt-3.5-turbo while experimenting since it will yield the best results''}.

Concluding, we expect that with careful prompt engineering, one can squeeze out some additional performance points, but we do not expect that this will lead to contradicting findings, e.g., that ChatGPT in a zero-shot prompting setting is competitive to fine-tuned models.

\section{Conclusions}

In conclusion, we evaluated the performance of the latest OpenAI's ChatGPT model on the LexGLUE benchmark in zero-shot and few-shot settings feeding examples in a templated instruction-following format. Our evaluation shows that ChatGPT achieves an average micro-F1 score of 49.0\% across LexGLUE tasks in a zero-shot setting, significantly exceeding random guess baselines, while outperformed by smaller fine-tuned models by approx. 30\%. Furthermore, in some cases, such as the ECtHR B and LEDGAR datasets, the performance of the model is very impressive, achieving micro-F1 scores of 62.8\% and 70.1\%, respectively.

Overall, our findings suggest that ChatGPT has overall poor performance in legal text classification tasks when provided with instruction-following inputs in a zero-shot setting, and hence should not be expected to be adequate in production out-of-the-box in similar settings (similar tasks with fixed/pre-defined label sets). However, further research is needed to explore the limits of GPT-3.5 models considering more delicate prompt engineering and more flexible evaluation. Moreover, OpenAI's GPT models are generic, and we can highly expect that similarly sized legal-oriented GPT models pre-trained on legal corpora and more similar task instructions would achieve improved performance.

\section*{Acknowledgments}
This work is funded by the Innovation Fund Denmark (IFD)\footnote{\url{https://innovationsfonden.dk/en}} under File No.\ 0175-00011A.

\bibliography{anthology,acl}
\bibliographystyle{acl_natbib}
\appendix
\end{document}